\documentclass[letterpaper, 10 pt, conference]{ieeeconf}  

\IEEEoverridecommandlockouts                              

\usepackage{graphicx}
\usepackage{epstopdf}
\usepackage{subcaption}
\usepackage{color}
\usepackage{multirow}
\usepackage{adjustbox}
\usepackage{booktabs}
\usepackage[table,xcdraw]{xcolor}
\usepackage{caption}
\usepackage{amsmath}
\usepackage{url}
\usepackage{array} 
\usepackage{graphicx}
\usepackage{amsmath}
\usepackage{amssymb}
\usepackage{booktabs}
\usepackage{cite}
\usepackage{bbding}

\definecolor{barrier}{rgb}{1, 0.47058824, 0.19607843}
\definecolor{bicycle}{rgb}{1, 0.75294118, 0.79607843}
\definecolor{bus}{rgb}{1, 1, 0.0}
\definecolor{car}{rgb}{0.0, 0.58823529, 0.96078431}
\definecolor{construction}{rgb}{0, 1, 1}
\definecolor{motorcycle}{rgb}{1, 0.49803922, 0}
\definecolor{pedestrian}{rgb}{1, 0, 0}
\definecolor{cone}{rgb}{1, 0.94117647, 0.58823529}
\definecolor{trailer}{rgb}{0.52941176, 0.23529412, 0}
\definecolor{truck}{rgb}{0.62745098, 0.1254902, 0.94117647}

\definecolor{driveable}{rgb}{1, 0, 1}
\definecolor{flat}{rgb}{0.54509804,0.5372549,0.5372549}
\definecolor{sidewalk}{rgb}{0.29411765,0,0.29411765}
\definecolor{terrain}{rgb}{0.58823529,0.94117647,0.31372549}
\definecolor{manmade}{rgb}{0.90196078,0.90196078,0.98039216}
\definecolor{vegetation}{rgb}{0,0.68627451,0}
\definecolor{ego_vehicle}{rgb}{0,0,0}

\title{\LARGE \bf
InverseMatrixVT3D: An Efficient Projection Matrix-Based Approach for 3D Occupancy Prediction
}

\author{Zhenxing Ming, Julie Stephany Berrio,  Mao Shan, Stewart Worrall
\thanks{The authors are with the Australian Centre for Robotics (ACFR) at the University of Sydney (NSW, Australia). E-mails: {\tt\small{\{d.ming, j.berrio, m.shan,
s.worrall}\}@acfr.usyd.edu.au}.}%
}

\begin{document}

\maketitle
\thispagestyle{empty}
\pagestyle{empty}

\begin{abstract}

This paper introduces InverseMatrixVT3D, an efficient method for transforming multi-view image features into 3D feature volumes for 3D semantic occupancy prediction. Existing methods for constructing 3D volumes often rely on depth estimation, device-specific operators, or transformer queries, which hinders the widespread adoption of 3D occupancy models. In contrast, our approach leverages two projection matrices to store the static mapping relationships and matrix multiplications to efficiently generate global Bird's Eye View (BEV) features and local 3D feature volumes. Specifically, we achieve this by performing matrix multiplications between multi-view image feature maps and two sparse projection matrices. We introduce a sparse matrix handling technique for the projection matrices to optimize GPU memory usage.
Moreover, a global-local attention fusion module is proposed to integrate the global BEV features with the local 3D feature volumes to obtain the final 3D volume. We also employ a multi-scale supervision mechanism to enhance performance further. Extensive experiments performed on the nuScenes and SemanticKITTI datasets reveal that our approach not only stands out for its simplicity and effectiveness but also achieves the top performance in detecting vulnerable road users (VRU), crucial for autonomous driving and road safety. The code has been made available at: https://github.com/DanielMing123/InverseMatrixVT3D

\end{abstract}

\section{INTRODUCTION}

Understanding the surrounding scene's three-dimensional (3D) geometry is fundamental to developing autonomous driving (AV) systems. While lidar-based methods that utilize explicit depth measurements have been performing exceptionally well on public datasets \cite{nuscenes,semantickitti}, they are hindered by the expensive cost of sensors and the sparsity of data points. As a result, the broader application of lidar-based methods is limited.

Vision-centric AV systems have garnered significant attention in recent years as a promising strategy due to its cost-effectiveness, stability, and generality. By utilizing multi-camera images as inputs, this approach has demonstrated competitive performance across various 3D perception tasks, including depth estimation \cite{surrounddepth,r3d3}, 3D object detection \cite{bevdet,bevdepth,bevformer}, online high-definition (HD) map construction \cite{maptr,maptrv2,hdmapnet}, and semantic map construction \cite{fiery,stretchbev}.

\begin{figure}[t] 
\centering 
\includegraphics[width=0.94\columnwidth]{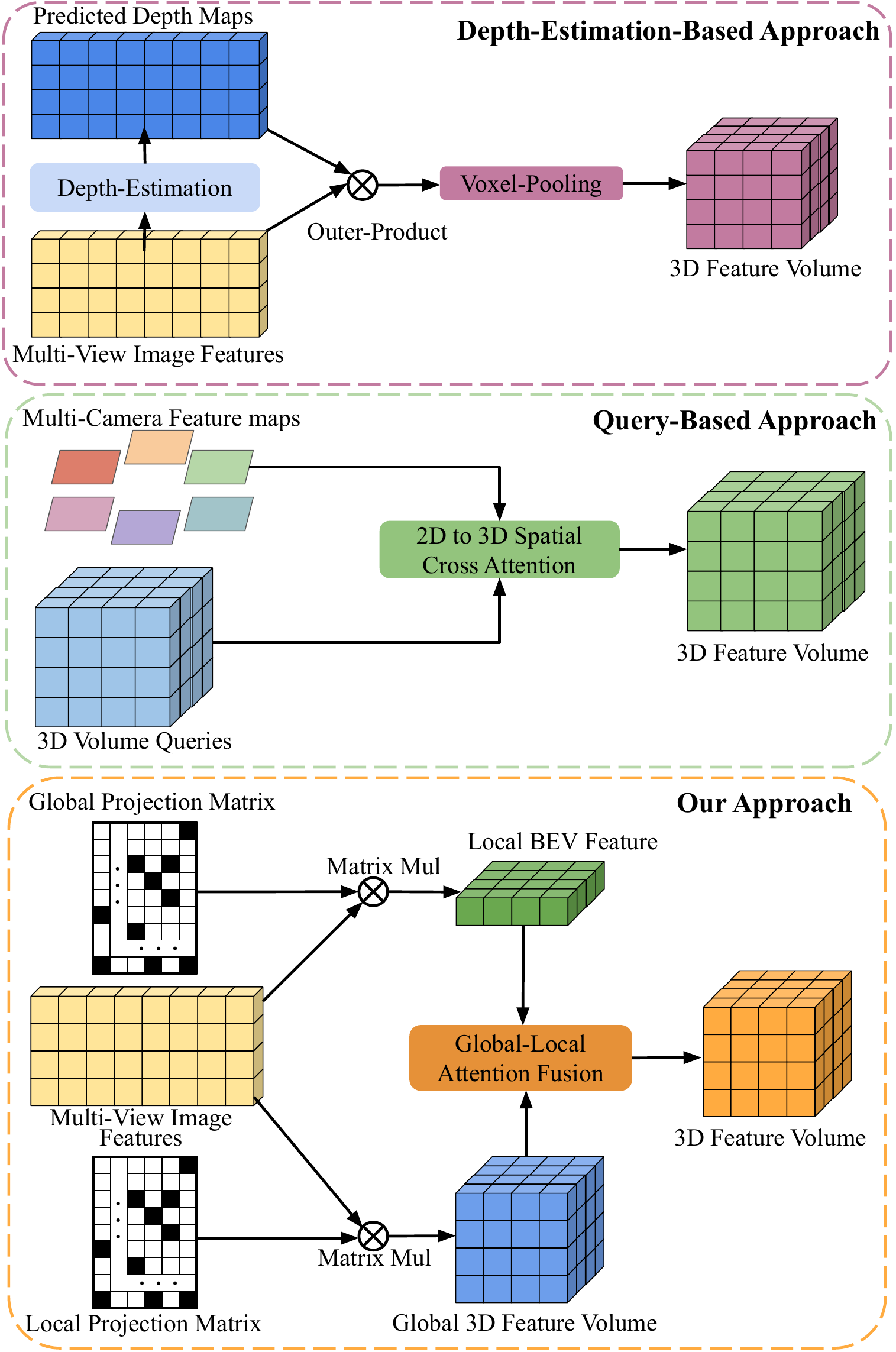} 
\caption{\small Pipeline of three approaches: Depth-Estimation-based approach (upper), Query-based approach (middle), and our approach (lower). We simplify the generation of the 3D Feature Volume by adopting a matrix multiplication method.} 
\label{Intro} 
\end{figure}

3D object detection based on fusing surround view cameras can be crucial in 3D perception. However, it faces challenges in handling new scenarios. One of the challenges is the finite number of semantic classes in the training dataset, making it difficult to create a model for every potential scenario that might be encountered on the road. In contrast, a more practical approach to depict the vehicle's surrounding environment is by directly reconstructing the 3D scenes.
In pursuit of this objective, several methods \cite{tpvformer,surroundocc,occformer} have been investigated to predict the 3D occupancy of a scene directly. These methods involve voxelizing the 3D space and assigning a probability to each voxel to determine its occupancy state--whether it is occupied or not. We argue that 3D occupancy serves as an adequate representation of the vehicle's surrounding environment. This representation inherently ensures geometric consistency and can accurately describe occluded areas. Furthermore, it is more robust towards object classes that do not exist in the training dataset. Despite the promise of these methods, their inner structure is quite complex, and some methods require additional sensors to provide supervision signals. For instance, the method proposed in \cite{occformer} relies on a lidar sensor to enhance performance through depth estimation supervision (Fig. \ref{Intro}, upper). Additionally, the methods proposed in \cite{tpvformer,surroundocc} extensively employ query-based modules \cite{deformabledetr} to aggregate image features for the final 3D feature volume (Fig. \ref{Intro}, middle).

To efficiently and effectively represent a 3D scene using 3D occupancy, we propose InverseMatrixVT3D (Fig. \ref{Intro}, bottom). Our method focuses on constructing projection matrices and simplifying the generation of local 3D feature volumes and global Bird's Eye View (BEV) features through matrix multiplication between multi-scale feature maps and projection matrices. Additionally, we employ a sparse matrix handling technique to optimize GPU memory usage when using these sparse projection matrices. Furthermore, we introduce a global-local fusion module to integrate global BEV features with local 3D feature volumes, resulting in the final 3D volume. We also apply a multi-scale supervision mechanism to each level to further enhance performance. Through comparisons with other state-of-the-art (SOTA) algorithms on the nuScenes and SemanticKITTI benchmarks, we demonstrate our method not only excels in its simplicity and effectiveness but also achieves the best performance in detecting vulnerable road users (VRU), i.e. pedestrians, motorcycles, and bicycles, which is a critical task for autonomous driving and road safety.  

The main contributions of this paper are summarized below. 
\begin{itemize}
    \item A novel projection matrix-based approach is proposed to simplify the local 3D feature volume and global BEV feature construction.
    \item A global-local fusion module is proposed to integrate global long-range information from the BEV feature with local spatial detail information from the 3D feature volume, resulting in the final 3D volume.
    \item We compare our approach with other state-of-the-art (SOTA) algorithms in the 3D semantic occupancy prediction task to prove the simplicity and effectiveness of our method.
\end{itemize}

The remainder of this paper is structured as follows: Section \ref{literature} provides an overview of related research and identifies the key differences between this study and previous publications. Section \ref{model} outlines the general framework of InverseMatrixVT3D and offers a detailed explanation of the implementation of each module. Section \ref{simulation} presents the findings of our experiments. Finally, Section \ref{conclusion} provides the conclusion of our work.

\section{Related Work}\label{literature}

\begin{figure*}[h]
\vspace{2mm}
     \centering
     \begin{subfigure}[]{\textwidth}
         \centering
         \includegraphics[width=\textwidth]{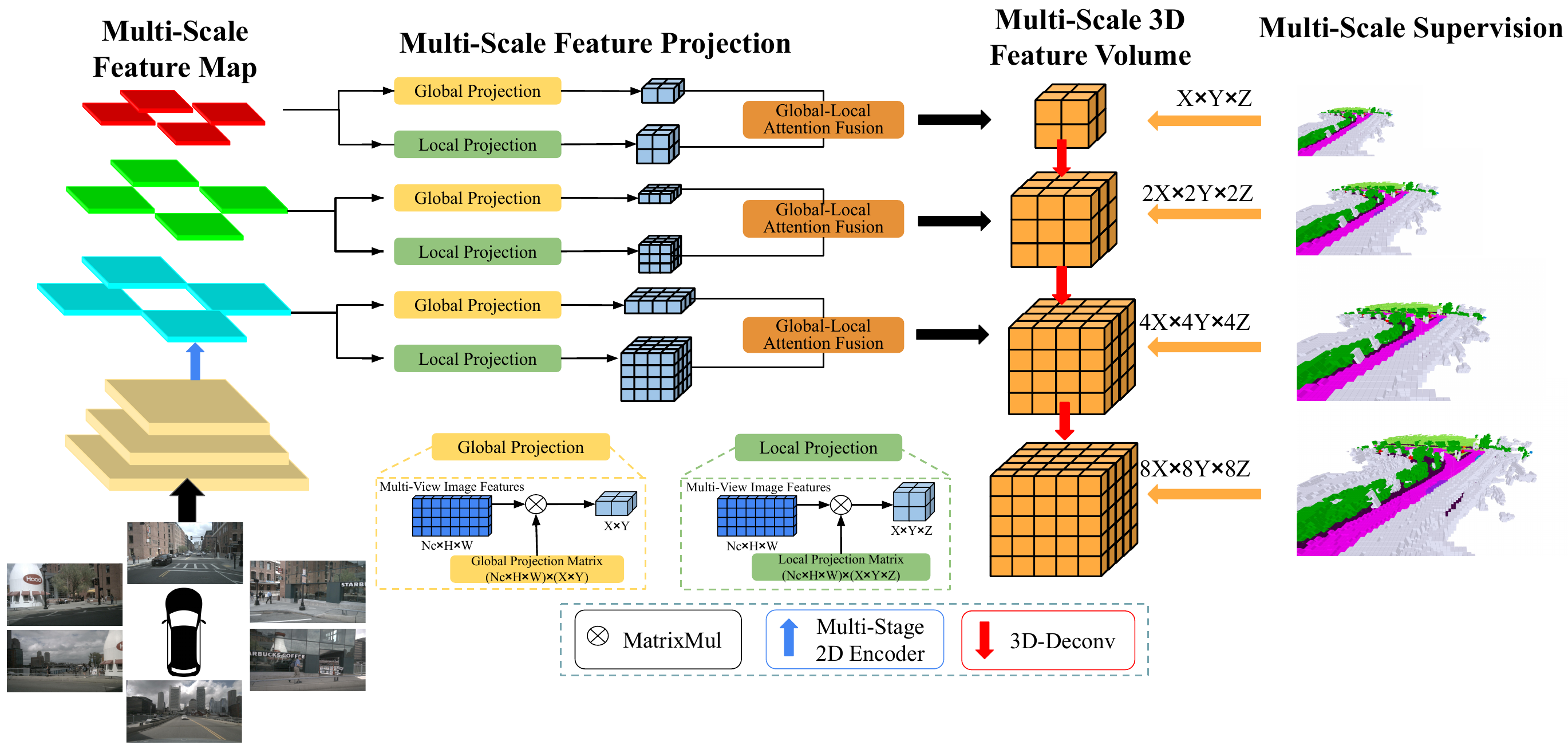}
     \end{subfigure}
        \caption{\small \textbf{Overall architecture of InverseMatrixVT3D.} Firstly, the multi-camera images were inputted into the 2D backbone network to extract features at multiple scales. Subsequently, a multi-scale global local projection module was employed to construct multi-scale 3D feature volumes and BEV planes. A global-local attention fusion module was applied to each 3D feature volume and BEV plane at every level to obtain the final 3D feature volume. Finally, the 3D volume at each level was upsampled using 3D deconvolution for skip-connection, and a supervision signal was also applied at each level.}
        \label{InverseMatrixVT3D}
\end{figure*}

\subsection{Depth-Estimation Based 3D Semantic Occupancy Prediction}
Based on the success of depth-estimation-based BEV perception algorithms, several works \cite{milo, multi, fbocc, occformer} have focused on constructing the 3D feature volume using pseudo-3D points. These approaches replace the previous splat operation in LSS \cite{LSS}, which generates the BEV feature, with a voxel-pooling operation. This new approach voxelizes the pseudo-3D point cloud and proposes several refinement modules to enhance the 3D feature volume.
OCCFormer \cite{occformer} introduces a dual-path transformer block to refine the BEV slice of the 3D feature volume, enhancing the long-range modeling capability of their model. FB-OCC \cite{fbocc} proposes an additional backward view-transformation module to improve the semantic information in the final 3D feature volume. Multi-Scale Occ \cite{multi} leverages a multi-scale fusion mechanism to capture global and local detail information in the 3D feature volume.

While depth-estimation-based approaches have achieved remarkable performance, they have one significant drawback: requiring depth ground truth labels to boost depth-estimation performance, boosting the model's overall performance. This requirement introduces extra effort during the training process. In this paper, we propose a method that eliminates the need for depth estimation by solely relying on multi-view images to construct a 3D feature volume. Our approach achieves superior performance compared to depth-estimation-based approaches.


\subsection{Query-Based 3D Semantic Occupancy Prediction}
Building upon the success of query-based BEV perception algorithms, TPVFormer \cite{tpvformer} introduces an extension to the BEV query by encompassing three perpendicular planes. This approach aims to capture the 3D world from multiple orthogonal perspectives. Similarly, SurroundOcc \cite{surroundocc} further expands the three-perpendicular plane concept to a 3D query volume. Leveraging intrinsic and extrinsic parameters, each query vector is projected onto multi-view images to aggregate dense features. Additionally, PanoOcc \cite{panoocc} proposes an efficient method for processing the dense 3D feature volume using a sparse representation approach.

Given the success of query-based approaches, it is important to highlight that the extensive use of transformer blocks in these methods often results in slow and inefficient training processes and high GPU memory consumption. In contrast, our proposed method eliminates the requirement for transformer-based querying and depth estimation while constructing the 3D feature volume. Our approach significantly improves model efficiency and enhances overall performance.

\section{InverseMatrixVT3D}\label{model}
In this paper, our objective is to generate a dense 3D occupancy grid of the surrounding scene using multi-camera images $Img=\left \{Img^{1},Img^{2},\cdots,Img^{N} \right \}$. Formally, the problem can be described as follows:
\begin{equation}
    3DOcc = VT(Img^{1},Img^{2},\dots,Img^{N})
\end{equation}
where VT is the neural network that leverages the projection matrix to aggregate features for 3D occupancy, the final 3D occupancy prediction result denoted as $3DOcc \in R^{X\times Y \times Z}$, represents the semantic property of the grids and has values ranging from 0 to 16. In our case, a class value of 0 indicates that the grid is empty.


Fig. \ref{InverseMatrixVT3D} exhibits the overall architecture of our method. Initially, given a set of surrounding multi-camera images, we first use a 2D backbone network(e.g.ResNet101-DCN) to extract $Nc$ cameras and $L$ levels multi-scale features $X=\left \{ \left \{ X_{n}^{l}  \right \}_{n=1}^{Nc}\in R^{C_{l} \times H_{l} \times W_{l}} \right \} _{l=1}^{L}$. For each level, we construct two projection matrices, namely, global projection matrix $VT\_XY^{l} \in R^{(Nc\times H_{l} \times W_{l}) \times (X_{l} \times Y_{l})}$ and local projection matrix $VT\_XYZ^{l} \in R^{(Nc\times H_{l} \times W_{l}) \times (X_{l} \times Y_{l} \times Z_{l})}$. The feature maps at each level are multiplied with these projection matrices, resulting in the 3D feature volume $F_{local}^{l}\in \mathbb{R}^{C_l \times X_l \times Y_l \times Z_l}$ and the Bird's-eye view (BEV) feature $F_{global}^{l}\in \mathbb{R}^{C_l \times X_l \times Y_l}$. Subsequently, the global-local attention fusion module merges information from these two features, producing the final 3D volume. Furthermore, the 3D volume at each level is upsampled using 3D deconvolution and integrated with the higher-level 3D volume through skip-connection. Finally, the dense occupancy ground truth from \cite{surroundocc} is applied to supervise the 3D volume at each level with a decayed loss weight.

\subsection{Multi-Camera Images Feature Extractor}
The purpose of the multi-camera image feature extractor is to capture both spatial and semantic features from the surrounding perspective view. These extracted features are the basis for the subsequent 3D occupancy prediction task. In our approach, we first employ a 2D backbone network to extract multi-scale feature maps. Then, a feature-pyramid network (FPN) is followed to further fuse feature output from different stages of the 2D backbone. The resulting feature maps have resolutions that are $\frac{1}{8}$, $\frac{1}{16}$, and $\frac{1}{32}$ of the input image resolution, respectively. The feature maps with smaller resolutions contain abundant semantic information, which assists the model in predicting the semantic class of each voxel grid. Conversely, the feature maps with larger resolutions provide richer spatial details and better guide the model regarding whether the current voxel grid is occupied or unoccupied.

\subsection{Global and Local Projection Matrix Generation}

In our approach, constructing global and local projection matrices is critical in gathering information for the local 3D feature volume and the global BEV feature. This differs from the method used in Occformer\cite{occformer}, which relies on depth estimation and voxel pooling, or the methods employed in TPVFormer\cite{tpvformer} and SurroundOcc\cite{surroundocc}, which involve transformer-based query.

\begin{figure}[h] 
\centering 
\includegraphics[width=0.85\columnwidth]{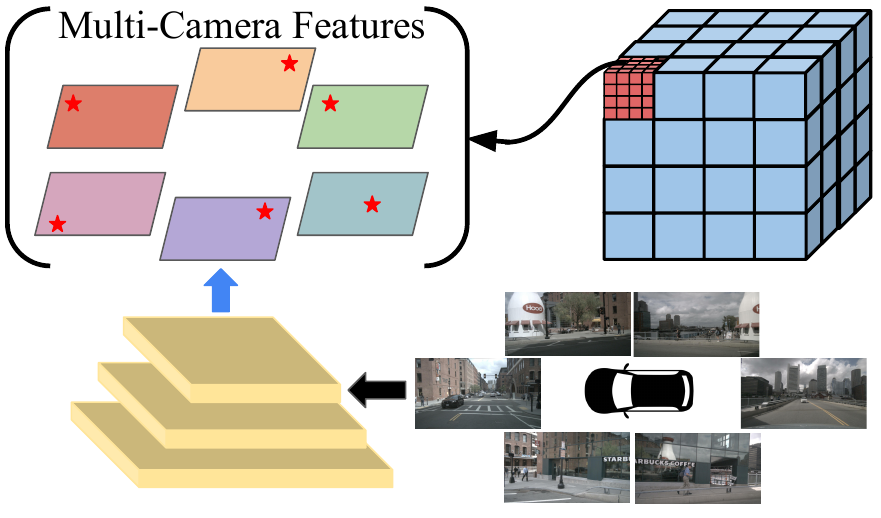} 
\caption{\small The predefined 3D volume feature sampling process. Each voxel grid is initially divided into $N^{3}$ subspaces along the vertical and horizontal directions, and the center of each subspace serves as a sample point. These sample points are then projected onto all the multi-view feature maps to aggregate the corresponding features for the voxel grid.} 
\label{Sample} 
\end{figure}

To begin with, we establish a set of predefined 3D volume spaces denoted by $F_{3D}^{l}\in \mathbb{R}^{X^{l}\times Y^{l}\times Z^{l}}$ for each level of multi-view feature maps under the ego vehicle's coordinate system. The ego vehicle is positioned at the center of these 3D volumes. Within each 3D volume, we divide the voxel grid equally into $N^3$ subspaces along the horizontal and vertical directions. Each center of the subspace serves as a sample point. Thus, for each level of the 3D volume space, we have a total of $X^{l}\times Y^{l}\times Z^{l}\times N^{3}$ sample points. Next, we project each sample point from each voxel grid onto the corresponding level of multi-view feature map $X^{l}$ using extrinsic and intrinsic parameters. Sample points that fall outside the boundaries of the feature maps or generate negative depth values are filtered out. The features hit by each sample point are then aggregated, resulting in a feature vector representing the corresponding voxel grid (see Fig. \ref{Sample}).

\begin{figure}
    \centering
    \begin{subfigure}[]{0.49\columnwidth}
    \centering
        \includegraphics[width=\columnwidth]{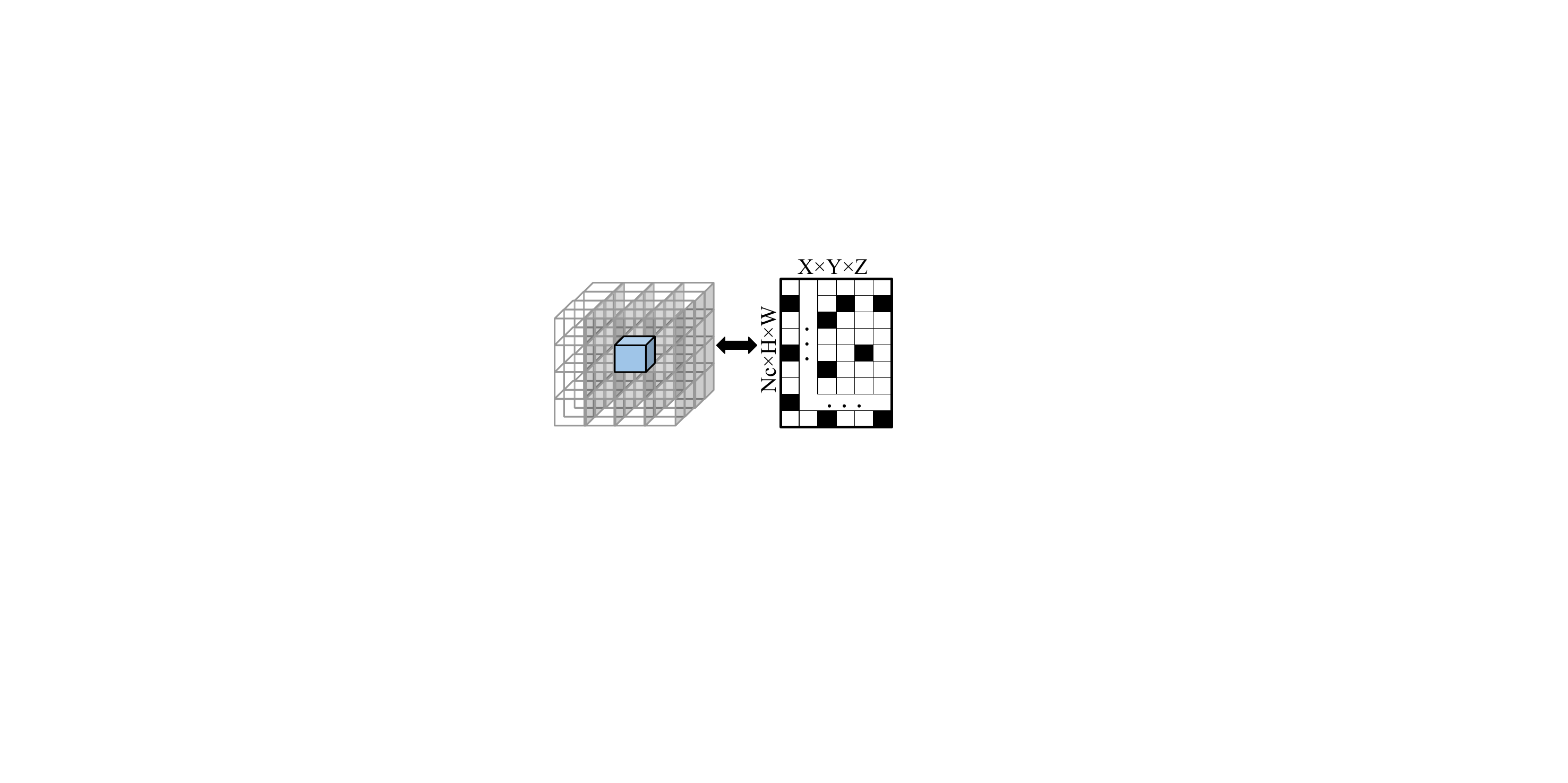}
        \caption{Local projection matrix}
        \label{VT_XYZ} 
    \end{subfigure}
    \begin{subfigure}[]{0.49\columnwidth}
    \centering
        \includegraphics[width=\columnwidth]{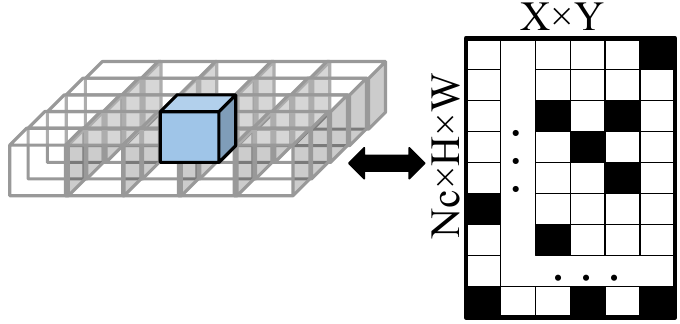}
        \caption{Global projection matrix}
        \label{VT_XY} 
    \end{subfigure}
    \caption{\small (a) The local projection matrix VT\_XYZ represents the static mapping relationship for the feature sampling process of the local 3D feature volume. (b) The global projection matrix VT\_XY represents the static mapping relationship for the feature sampling process of the global Bird's Eye View (BEV) feature.}
    \label{fig:enter-label}
\end{figure}

The feature sampling process is a static mapping and can be represented by constructing the projection matrices $VT\_XYZ^{l} \in R^{(Nc\times H_{l} \times W_{l}) \times (X_{l} \times Y_{l} \times Z_{l})}$ as exhibited in Fig. \ref{VT_XYZ}. Moreover, height information can be compressed further by constructing the projection matrix $VT\_XY^{l} \in R^{(Nc\times H_{l} \times W_{l}) \times (X_{l} \times Y_{l})}$ as shown in Fig. \ref{VT_XY}. As a result, the generation of both the local 3D feature volume and global BEV feature can be greatly simplified as a matrix multiplication between the multi-view feature maps and the two projection matrices:
\begin{equation}
    F_{local}^{l}= X^{l} \cdot VT\_XYZ^{l}  
\end{equation}
and
\begin{equation}
    F_{global}^{l}= X^{l} \cdot VT\_XY^{l}  
\end{equation}
where $X^{l}\in R^{N_{c}\times H_{l}\times W_{l}}$ represents the multi-view feature maps at the $l$-th level.

During the construction process of the global and local projection matrices, we observed that these matrices exhibit extensive sparsity. Consequently, the GPU memory utilization for constructing these matrices increases exponentially with their resolution. To optimize GPU memory utilization, we utilize the compressed sparse row (CSR) technique \cite {csr}. This technique stores only the non-zero values and their associated indices when constructing and storing the sparse matrices. By applying this technique, we can dramatically decrease the GPU memory usage for our highest 3D volume resolution from 15GB to 200MB.

\subsection{Global Local Attention Fusion}
\begin{figure}[htpb] 
\centering 
\includegraphics[width=0.48\textwidth]{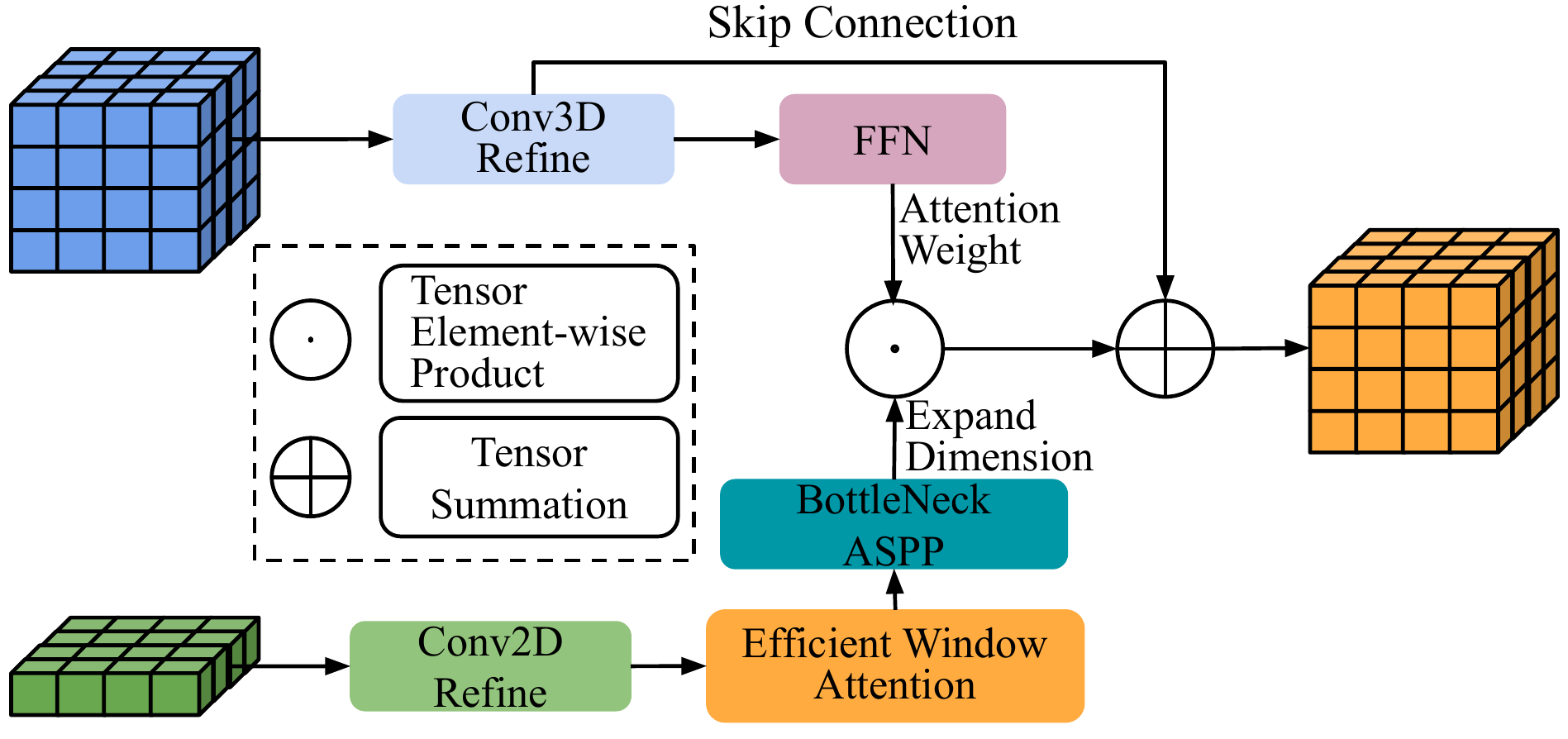} 
\caption{\small The global-local fusion module. The global BEV feature and the local 3D feature volume are refined using traditional convolutional layers. The global BEV feature undergoes additional enhancement through an efficient window attention module and a bottleneck ASPP module. It then merges with the local 3D feature volume, resulting in the final 3D volume.} 
\label{Global_Local} 
\end{figure}

To enhance the ability of the final 3D feature volume to capture both global and local details, we introduce the Global-Local Attention Fusion module. The detailed structure of the Global-Local Attention Fusion module is depicted in Fig. \ref{Global_Local}. 

Inspired by \cite{squeeze}, the module begins by applying traditional 2D and 3D convolution operations to enhance the global BEV feature and the local 3D feature volume. Building on recent advancements that emphasize the significance of locality and efficiency in transformers and the importance of the BEV plane \cite{occformer}, we incorporate an efficient window attention module from \cite{efficientvit} and a bottleneck ASPP module to further refine the global BEV feature. Simultaneously, the local 3D feature volume is processed through the Feed-Forward Network (FFN) to generate attention weights. The BEV feature, refined by the window attention and bottleneck ASPP module, undergoes dimension expansion and is element-wise multiplied by the attention weights. This resulting feature is then added to the refined local 3D feature volume to produce the final 3D feature volume. The whole process can be described as follows:
\begin{equation}
    F_{3DOcc}^{l}=F_{local}^{l} + \sigma(FNN(F_{local}^{l}))\cdot Expand(F_{global}^{l})  
\end{equation}
where $\sigma()$ refers to applying the sigmoid function to the output of the FFN; this function constrains the attention weight to the [0,1] range. Additionally, the Expand operation refers to expanding the dimension of $F_{global}^{l}$ to match the dimension of $F_{local}^{l}$.

\begin{table*}[t]
  \centering
  \begin{adjustbox}{width=\textwidth}
  \begin{tabular}{c|c|c|c|cc|cccccccccccccccc}
    \toprule
    Method & Backbone & Params & Resolution & IoU & mIoU & \rotatebox{90}{\textcolor{barrier}{$\bullet$} barrier} & \rotatebox{90}{\textcolor{bicycle}{$\bullet$} bicycle} & \rotatebox{90}{\textcolor{bus}{$\bullet$} bus} & \rotatebox{90}{\textcolor{car}{$\bullet$} car} & \rotatebox{90}{\textcolor{construction}{$\bullet$} const. veh.} & \rotatebox{90}{\textcolor{motorcycle}{$\bullet$} motorcycle} & \rotatebox{90}{\textcolor{pedestrian}{$\bullet$} pedestrian} & \rotatebox{90}{\textcolor{cone}{$\bullet$} traffic cone} & \rotatebox{90}{\textcolor{trailer}{$\bullet$} trailer} & \rotatebox{90}{\textcolor{truck}{$\bullet$} truck} & \rotatebox{90}{\textcolor{driveable}{$\bullet$} drive. surf.} & \rotatebox{90}{\textcolor{flat}{$\bullet$} other flat} & \rotatebox{90}{\textcolor{sidewalk}{$\bullet$} sidewalk} & \rotatebox{90}{\textcolor{terrain}{$\bullet$} terrain} & \rotatebox{90}{\textcolor{manmade}{$\bullet$} manmade} & \rotatebox{90}{\textcolor{vegetation}{$\bullet$} vegetation} \\
     \midrule 
    MonoScene\cite{monoscene} & ResNet101-DCN & - & $200\times200\times16$ & 23.96 & 7.31 & 4.03 & 0.35 & 8.00 & 8.04 & 2.90 & 0.28 & 1.16 & 0.67 & 4.01 & 4.35 & 27.72 & 5.20 & 15.13 & 11.29 & 9.03 & 14.86\\
    Atlas* \cite{atlas} & - & - & $200\times200\times16$ & 28.66 & 15.00 & 10.64 & 5.68 & 19.66 & 24.94 & 8.90 & 8.84 & 6.47 & 3.28 & 10.42 & 16.21 & 34.86 & 15.46 & 21.89 & 20.95 & 11.21 & 20.54 \\
    BEVFormer*\cite{bevformer} & ResNet101-DCN & 59M & $200\times200$ & 30.50 & 16.75 & 14.22 & 6.58 & 23.46 & 28.28 & 8.66 & 10.77 & 6.64 & 4.05 & 11.20 & 17.78 & 37.28 & 18.00 & 22.88 & 22.17 & 13.80 & 22.21 \\
    TPVFormer\cite{tpvformer} & ResNet101-DCN & 69M & $200\times200\times16$ & 11.51 & 11.66 & 16.14 & 7.17 & 22.63 & 17.13 & 8.83 & 11.39 & 10.46 & 8.23 & 9.43 & 17.02 & 8.07 & 13.64 & 13.85 & 10.34 & 4.90 & 7.37 \\
    TPVFormer* & ResNet101-DCN & 69M & $200\times200\times16$ & 30.86 & 17.10 & 15.96 & 5.31 & 23.86 & 27.32 & 9.79 & 8.74 & 7.09 & 5.20 & 10.97 & 19.22 & 38.87 & 21.25 & 24.26 & 23.15 & 11.73 & 20.81 \\
    C-CONet*\cite{openoccupancy} & ResNet101 & 118M & $200\times200\times16$ & 26.10 & 18.40 & 18.60 & 10.00 & 26.40 & 27.40 & 8.60 & 15.70 & 13.30 & 9.70 & 10.90 & 20.20 & 33.00 & 20.70 & 21.40 & 21.80 & 14.70 & 21.30 \\
    LMSCNet*\cite{lmscnet} & - & - & $200\times200\times16$ & 36.60 & 14.90 & 13.10 & 4.50 & 14.70 & 22.10 & 12.60 & 4.20 & 7.20 & 7.10 & 12.20 & 11.50 & 26.30 & 14.30 & 21.10 & 15.20 & 18.50 & 34.20 \\
    L-CONet*\cite{openoccupancy} & - &  & $200\times200\times16$ & \textbf{39.40} & 17.70 & 19.20 & 4.00 & 15.10 & 26.90 & 6.20 & 3.80 & 6.80 & 6.00 & 14.10 & 13.10 & 39.70 & 19.10 & 24.00 & 23.90 & 25.10 & 35.70 \\
    SurroundOcc*\cite{surroundocc} & ResNet101-DCN & 180M & $200\times200\times16$ & 31.49 & \textbf{20.30} & 20.59 & 11.68 & 28.06 & 30.86 & 10.70 & 15.14 & 14.09 & 12.06 & 14.38 & 22.26 & 37.29 & 23.70 & 24.49 & 22.77 & 14.89 & 21.86 \\
    \hline
    
    InverseMatrixVT3D* (Base) & ResNet101-DCN & \textbf{67M} & $200\times200\times16$ & 31.85 & 18.88 & 18.39 & 12.46 & 26.30 & 29.11 & 11.00 & 15.74 & \textbf{14.78} & 11.38 & 13.31 & 21.61 & 36.30 & 19.97 & 21.26 & 20.43 & 11.49 & 18.47 \\
    InverseMatrixVT3D* (Post-Fix) & ResNet101-DCN & \textbf{67M} & $200\times200\times16$ & 26.79 & 15.81 & 16.47 & 10.27 & 21.28 & 28.29 & 8.32 & 13.29 & 12.90 & 8.41 & 10.96 & 18.49 & 32.43 & 11.79 & 18.27 & 15.45 & 9.87 & 16.52 \\
    InverseMatrixVT3D* (Pre-Fix) & ResNet101-DCN & \textbf{67M} & $200\times200\times16$ & 31.30 & 18.42 & 18.06 & \textbf{12.72} & 25.99 & 28.00 & 10.15 & \textbf{15.98} & 14.31 & 10.61 & 12.49 & 20.58 & 35.61 & 19.40 & 21.00 & 20.30 & 11.06 & 18.59 \\
    \bottomrule
  \end{tabular}
  \end{adjustbox}
  \caption{\textbf{3D semantic occupancy prediction results on nuScenes validation set}. * means method is trained with dense occupancy labels from \cite{surroundocc}.}
  \label{occ}
\end{table*}

\begin{table*}[t]
  \centering
  \begin{adjustbox}{width=\textwidth}
  \begin{tabular}{c|cc|ccccccccccccccccccc}
    \toprule
    Method & IoU & mIoU & \rotatebox{90}{road} & \rotatebox{90}{sidewalk} & \rotatebox{90}{parking} & \rotatebox{90}{other-ground} & \rotatebox{90}{building} & \rotatebox{90}{car} & \rotatebox{90}{truck} & \rotatebox{90}{bicycle} & \rotatebox{90}{motorcycle} & \rotatebox{90}{other-vehicle} & \rotatebox{90}{vegetation} & \rotatebox{90}{trunk} & \rotatebox{90}{terrain} & \rotatebox{90}{person} & \rotatebox{90}{bicyclist} & \rotatebox{90}{motorcyclist} & \rotatebox{90}{fence} & \rotatebox{90}{pole} & \rotatebox{90}{traf.sign} \\
     \midrule 
    LMSCNet \cite{lmscnet} & 28.61 & 6.70 & 40.68 & 18.22 & 4.38 & 0.00 & 10.31 & 18.33 & 0.00 & 0.00 & 0.00 & 0.00 & 13.66 & 0.02 & 20.54 & 0.00 & 0.00 & 0.00 & 1.21 & 0.00 & 0.00 \\
    AICNet \cite{anisotropic} & 29.59 & 8.31 & 43.55 & 20.55 & 11.97 & 0.07 & 12.94 & 14.71 & 4.53 & 0.00 & 0.00 & 0.00 & 15.37 & 2.90 & 28.71 & 0.00 & 0.00 & 0.00 & 2.52 & 0.06 & 0.00 \\
    3DSketch \cite{3dsketch} & 33.30 & 7.50 & 41.32 & 21.63 & 0.00 & 0.00 & 14.81 & 18.59 & 0.00 & 0.00 & 0.00 & 0.00 & \textbf{19.09} & 0.00 & 26.40 & 0.00 & 0.00 & 0.00 & 0.73 & 0.00 & 0.00 \\
    JS3C-Net \cite{js3cnet} & \textbf{38.98} & 10.31 & 50.49 & 23.74 & 11.94 & 0.07 & \textbf{15.03} & \textbf{24.65} & 4.41 & 0.00 & 0.00 & 6.15 & 18.11 & \textbf{4.33} & 26.86 & 0.67 & 0.27 & 0.00 & 3.94 & 3.77 & 1.45 \\
    MonoScene \cite{monoscene} & 36.86 & 11.08 & \textbf{56.52} & \textbf{26.72} & 14.27 & 0.46 & 14.09 & 23.26 & 6.98 & 0.61 & 0.45 & 1.48 & 17.89 & 2.81 & 29.64 & 1.86 & 1.20 & 0.00 & 5.84 & \textbf{4.14} & 2.25 \\
    TPVFormer \cite{tpvformer} & 35.61 & 11.36 & 56.50 & 25.87 & \textbf{20.60} & \textbf{0.85} & 13.88 & 23.81 & 8.08 & 0.36 & 0.05 & 4.35 & 16.92 & 2.26 & \textbf{30.38} & 0.51 & 0.89 & 0.00 & \textbf{5.94} & 3.14 & 1.52\\
    \hline
    InverseMatrixVT3D (Base) & 36.22 & \textbf{11.81} & 52.99 & 25.84 & 20.04 & 0.09 & 13.17 & 24.08 & \textbf{10.25} & \textbf{1.85} & \textbf{2.65} & \textbf{6.80} & 16.98 & 3.09 & 27.77 & \textbf{4.01} & \textbf{3.13} & 0.00 & 4.94 & 4.05 & \textbf{2.67} \\
    \bottomrule
  \end{tabular}
  \end{adjustbox}
  \caption{\textbf{3D semantic scene completion performance on SemanticKITTI validation set}.}
  \label{kitti}
\end{table*}

\section{Experimental Results}\label{simulation}

\subsection{Implementation Details}
The InverseMatrixVT3D uses ResNet101-DCN \cite{residual} as a 2D backbone with a checkpoint from FCOS3D \cite{fcos3d} to extract image features. The features of stage 1,2,3 of the backbone are fed to FPN \cite{fpn}, resulting in 3-level multi-scale image features. The network architecture comprises four levels ($L=4$), with no skip connection applied to the highest level. For the paths corresponding to levels 1, 2, and 3, we employ divided schemas of N=3, 4, and 5, respectively, to create sets of sample points. The AdamW optimizer with an initial learning rate of 5e-5 and weight decay of 0.01 is employed for optimization. The learning rate is decayed using a multi-step scheduler. Regarding data augmentation, random resize, rotation and flip operations are applied in the image space, following established practices for BEV-based 3D object detection \cite{bevdet,bevdepth,bevstereo,bevformer} and the compared methods \cite{tpvformer, surroundocc, occformer, monoscene}. 
The predicted occupancy has a resolution of $200 \times 200 \times 16$ for full-scale evaluation. The model is trained to utilize eight A10 GPUs with 24GB of memory, and it has been trained for 2 days.

\subsection{Loss Function}
To train the model, we use focal loss \cite{focal}, Lovasz-softmax loss \cite{lovasz} and scene-class affinity loss \cite{monoscene} to handle the significant sparsity of free space in the ground truth dataset. The final loss is composed of:
\begin{equation}
    Loss=L_{focal} + L_{lovasz} + L_{scal}^{geo} + L_{scal}^{sem}
\end{equation}

\subsection{Dataset}
The nuScenes dataset \cite{nuscenes}, a vast autonomous driving dataset, serves as the data source for our experiments. As the test set lacks semantic labels, we train our model on the training set and assess its performance using the validation set. We set the range for occupancy prediction as [-50, 50] meters for the X and Y axes and [-5, 3] meters for the Z axis. The input images have a $1600 \times 900$ pixels resolution, while the final output occupancy is represented with a resolution of $200 \times 200 \times 16$ for the base version. We have conducted experiments on 3D semantic occupancy prediction tasks to provide quantitative results. The dense labels used for the 3D semantic occupancy prediction task are sourced from \cite{surroundocc}. Additionally, we provide qualitative visualizations of the results for the 3D semantic occupancy prediction task.

To enhance the demonstration of the effectiveness of our approach, we conducted a monocular semantic scene completion experiment on the SemanticKITTI dataset \cite{semantickitti} employing the left RGB camera. SemanticKITTI contains annotated outdoor LiDAR scans with 21 semantic labels. The input image resolution is $1241\times376$, and the ground truth is voxelized into a grid of dimensions 256x256x32 with a voxel size of 0.2m. The evaluation of our model is performed on the validation set.

\subsection{Performance Evaluate Metrics}
To assess the performance of various state-of-the-art (SOTA) algorithms and compare them with our approach in the 3D semantic occupancy prediction task, we utilize the intersection over union (IoU) to evaluate each semantic class. Moreover, we employ the mean IoU overall semantic classes (mIoU) as a comprehensive evaluation metric:
\begin{equation}
    IoU=\frac{TP}{TP+FP+FN} 
\end{equation}
and
\begin{equation}
    mIoU=\frac{1}{Cls}\sum_{i=1}^{Cls}  \frac{TP_{i}}{TP_{i}+FP_{i}+FN_{i}} 
\end{equation}
where $TP$, $FP$, and $FN$ represent the counts of true positives, false positives, and false negatives in our predictions, respectively, while $Cls$ denotes the total class number.

\begin{figure*}[t]
\centering
\begin{subfigure}[]{\textwidth}
\includegraphics[width=\textwidth]{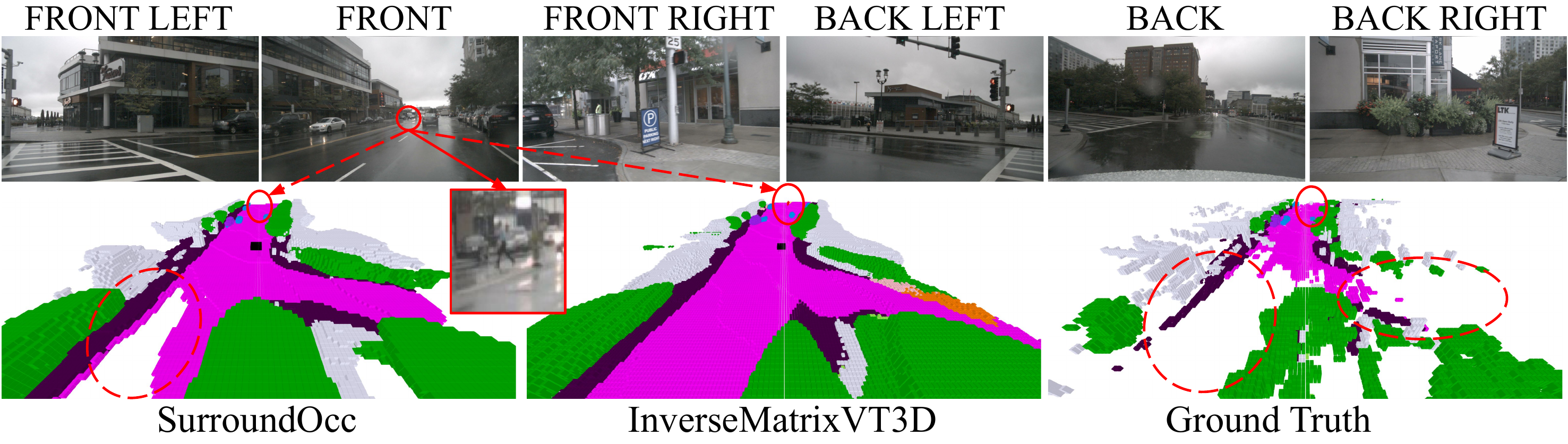}
\includegraphics[width=\textwidth]{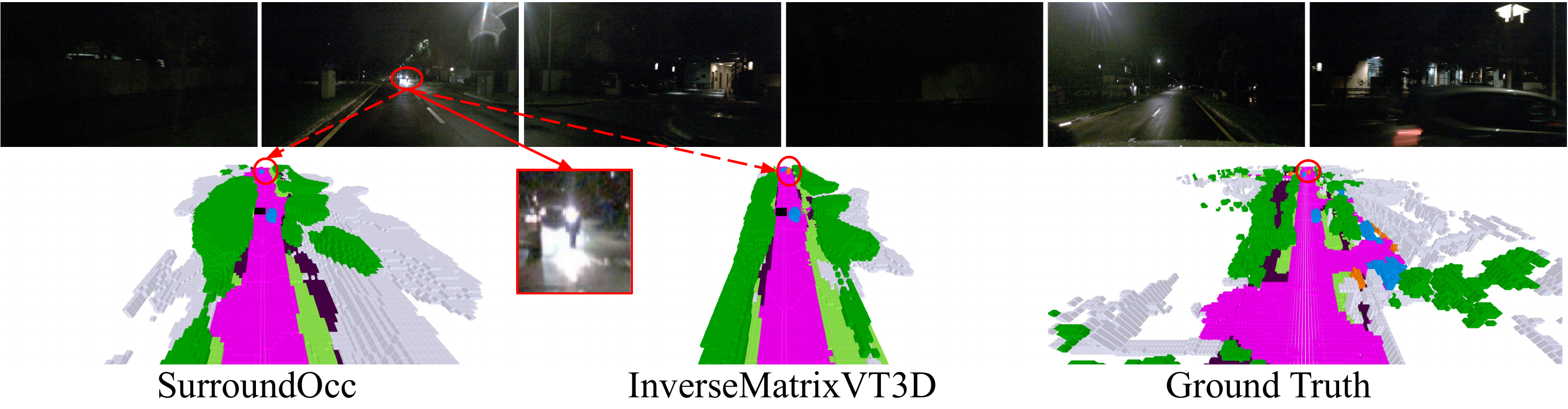}
\end{subfigure}
\textcolor{barrier}{$\bullet$}  barrier  \textcolor{bicycle}{$\bullet$} bicycle \textcolor{bus}{$\bullet$} bus \textcolor{car}{$\bullet$}  car  \textcolor{construction}{$\bullet$} construction vehicle \textcolor{motorcycle}{$\bullet$} motorcycle \textcolor{pedestrian}{$\bullet$} pedestrian \textcolor{cone}{$\bullet$} traffic cone \textcolor{trailer}{$\bullet$} trailer \textcolor{truck}{$\bullet$} truck
    
\textcolor{driveable}{$\bullet$} driveable surface \textcolor{flat}{$\bullet$} other flat \textcolor{sidewalk}{$\bullet$} sidewalk \textcolor{terrain}{$\bullet$} terrain \textcolor{manmade}{$\bullet$} manmade \textcolor{vegetation}{$\bullet$} vegetation \textcolor{ego_vehicle}{$\bullet$} ego vehicle \\

\caption{\small Challenging scenes qualitative analysis. Despite challenging lighting and weather conditions, our approach successfully predicts moving objects with high accuracy, even when they are far away from the ego vehicle. \textbf{Better viewed when zoomed in.}}
\label{qualitative}
\end{figure*}

\subsection{Model Performance Analysis}
We perform the task of multi-camera 3D semantic occupancy prediction on the nuScenes dataset and the monocular semantic scene completion task on the SemanticKITTI dataset. To evaluate the performance of our proposed InverseMatrixVT3D in each task, we compare it with other SOTA algorithms and present the results in Table \ref{occ} and \ref{kitti}, respectively. In Table \ref{occ}, our model has three different settings: the InverseMatrixVT3D (Base), where the projection matrices generation process is included during training and evaluation, the InverseMatrixVT3D (Post-Fix), which involves fixing the projection matrices after the model finished training for evaluation purposes, and the InverseMatrixVT3D (Pre-Fix), which fixed projection matrices in advance and directly incorporating them into the training process.

In the context of the 3D semantic occupancy prediction task using the nuScenes dataset, our model demonstrates very competitive performance. It outperforms several transformer-query-based methods and ranks second on the benchmark. Compared to transformer-query-based methods, our model exhibits the best performance in detecting  VRU on roadways, including bicycles, motorcycles, and pedestrians. Nevertheless, when dealing with background objects, like vegetation, manmade structures, and terrain, our model is inferior to transformer-query-based approaches. This may be attributed to the fact that many background objects are invisible, but due to the powerful generalization ability provided by the transformer, those transformer-query-based approaches can infer the background object's semantic label and its locations to some extent. Another point worth mentioning is that although our model failed to beat SurroundOcc, the size of our model only has a total of 67M trainable parameters, which is substantially smaller than SurroundOcc's 180M trainable parameters. Notably, the model's performance significantly diminishes when utilizing fixed projection matrices after training due to noticeable projection errors. The projection error can be alleviated by employing fixed projection matrices in advance and directly incorporating them into the training process. In the realm of the monocular semantic scene completion task on the SemanticKITTI dataset, our model exhibits similar characteristics to those observed in the 3D semantic occupancy prediction task and has delivered very competitive performance in comparison with other SOTA algorithms. These results highlight our model's superior 3D world modelling capability.

In general, our model aggregates features for the final 3D volume based on sampling at specific sampling locations. Compared to the transformer-based approach, whose sampling locations can be adjusted based on the query vector, our approach's sampling locations are fixed. Nevertheless, our method boasts a significantly higher sampling density than the transformer-based approach, allowing for dense feature aggregation. In essence, our model prioritizes sampling density over sampling flexibility, which leads to competitive performance and a much smaller model size.

\subsection{Challenging Scenes Qualitative Analysis}
We demonstrate our model's powerful 3D modelling capability and exceptional VRU detection performance by presenting the prediction results in challenging rainy and night-time scenes along with SurroundOcc prediction results and ground truth, as shown in Fig. \ref{qualitative}. Regarding the rainy scenario depicted in Fig. \ref{qualitative} upper, we observed an inconsistency in the ground truth labels. This discrepancy is attributed to the lidar's poor performance during rainy weather due to the reflection effect. Consequently, the ground truth labels become inconsistent. However, our model's prediction result is remarkably accurate. It not only fills in the missing information from the ground truth but also successfully detects a pedestrian crossing the road (indicated by the red circle), even though the person is located at a significant distance from the ego vehicle. Moreover, in the night-time scene illustrated in Fig. \ref{qualitative} bottom, our model effectively predicts the presence of a motorcycle (marked by the red circle) in the distance ahead despite the low ambient light conditions.

\subsection{Model Efficiency}
\begin{table}[htbp]
  \centering
  \begin{adjustbox}{width=0.9\columnwidth}
  \begin{tabular}{c|cc}
    \toprule
    {Method} & {Latency (s) ($\downarrow$)} & {Memory (GB) ($\downarrow$)} \\
    \midrule
    NeWCRFs \cite{newcrfs} & 1.07 & 14.5 \\
    MonoScene \cite{monoscene} & 0.87 & 20.3 \\
    Adabins \cite{adabins} & 0.75 & 15.5 \\
    SurroundDepth \cite{surrounddepth} & 0.73 & 12.4\\
    SurroundOcc \cite{surroundocc} & 0.34 & 5.9 \\
    TPVFormer \cite{tpvformer} & 0.32 & 5.1 \\
    BEVFormer \cite{bevformer} & \textbf{0.31} & \textbf{4.5} \\
    \hline
    InverseMatrixVT3D (Base) & 0.5 & 5.2 \\
    InverseMatrixVT3D (Fix) & 0.32 & 4.82 \\
    \bottomrule
  \end{tabular}
  \end{adjustbox}
  \caption{Model efficiency comparison of different methods. The experiments are performed on a single RTX 3090 using six multi-camera images. For input image resolution, all methods adopt $1600\times900$. $\downarrow$:the lower, the better.}
  \label{efficiency}
\end{table}
Table \ref{efficiency} compares the inference time and inference memory among different methods. The experiments are conducted on a single RTX 3090 using six multi-camera images. All methods adopt an image resolution of $1600\times900$. Our base model, which includes the projection matrices generation, runs 0.5 seconds for a single data sample, but if we fixed the projection matrices, remove the projection matrices generation procedure. Due to the utilization of projection matrices that facilitate the 3D volume generation, our fixed version model achieves exceptional real-time performance.

\subsection{Ablation Study}
\subsubsection{Global Local Attention Fusion}
We perform an ablation study on the global-local attention fusion module, and the experiment results are presented in Table \ref{abla_global_local}. The experimental results confirm that the BEV feature and its associated refinement procedures are crucial in improving model performance. Without the BEV feature, the final 3D volume is unable to capture long-range global semantic information, ultimately leading to performance degradation. 
\begin{table}[h]
  \centering
  \begin{adjustbox}{width=0.6\columnwidth}
  \begin{tabular}{ccc|c}
    \toprule
    BEV & Bott.ASPP & Eff.Win.Atten & mIoU$\uparrow$ \\
    \midrule
    \Checkmark & \Checkmark & \Checkmark & \textbf{14.44}\% \\
    \Checkmark & \Checkmark & \XSolidBrush & 9.04\% \\
    \Checkmark & \XSolidBrush & \Checkmark & 9.26\% \\
    \XSolidBrush & \XSolidBrush & \XSolidBrush & 8.45\% \\
    \bottomrule
  \end{tabular}
  \end{adjustbox}
  \caption{\small Ablation study for global-local attention fusion module. Bott.ASPP: bottleneck aspp, Eff.Win.Atten: efficient window attention. $\uparrow$:the higher, the better.}
  \label{abla_global_local}
\end{table}

\subsubsection{Multi-Scale Mechanism}
We conducted an ablation study on the multi-scale mechanism, and the experiment results are presented in Table \ref{abla_mul_scale}. The experimental results demonstrate that the multi-level supervision and coarse-to-fine refinement structure play a vital role during training. Without these two structures, our model experiences a performance degradation of at least 5\%. 
\begin{table}[h]
  \centering
  \begin{adjustbox}{width=0.6\columnwidth}
  \begin{tabular}{cc|c|c}
    \toprule
    Multi.Stru & Multi.Sup & Params & mIoU$\uparrow$ \\
    \midrule
    \Checkmark & \Checkmark & 67.18M & \textbf{14.44}\% \\
    \Checkmark & \XSolidBrush & 67.18M & 7.65\% \\
    \XSolidBrush & \XSolidBrush & 58.89M & 6.07\% \\
    \bottomrule
  \end{tabular}
  \end{adjustbox}
  \caption{\small Ablation study on multi-scale mechanism. Multi.Stru: multi-scale corse-to-fine refinement structure, Multi.Sup: multi-level supervision mechanism. $\uparrow$:the higher, the better.}
  \label{abla_mul_scale}
\end{table}

\subsubsection{Dividing Schemas For Each Level}
We performed an ablation study on the division schemas for each level, and the experiment results are presented in Table \ref{abla_divide}. We used different N settings for levels 1, 2, and 3 to generate sample points, and the experimental results aligned with our expectations. Specifically, when the divide setting N is larger, more sample points can be generated, leading to dense sampling and rich feature aggregation. We also observed a trend where a larger divide setting at higher levels significantly impacts the final model performance, indicating the increased importance of higher-level feature aggregation.
\begin{table}[h]
  \centering
  \begin{tabular}{ccc|c}
    \toprule
    level1 & level2 & level3 & mIoU$\uparrow$ \\
    \midrule
    3 & 4 & 5 & \textbf{14.44}\% \\
    2 & 3 & 4 & 8.45\% \\
    1 & 2 & 3 & 5.66\% \\
    1 & 4 & 6 & 7.73\% \\
    \bottomrule
  \end{tabular}
  \caption{\small Ablation study of dividing schemas for each level. $\uparrow$:the higher, the better.}
  \label{abla_divide}
\end{table}

\section{Conclusion}\label{conclusion}
In this paper, we propose InverseMatrixVT3D, a vision-centric 3D semantic occupancy prediction method. Our approach leverages predefined sample points for each scale of 3D volumes and constructs projection matrices to represent the fixed sampling process. Through matrix multiplication between multi-view feature maps and projection matrices in a multi-scale fashion, we generate local 3D feature volumes and global BEV features. These features are merged using our proposed global-local fusion module, resulting in the final 3D volume at each level. Lastly, the 3D volumes at each level are upsampled and fused using a 3D deconvolution layer. Unlike other SOTA algorithms, our approach does not require depth estimation or transformer-based query, making the 3D volume generation process simple and efficient. Extensive experiments conducted on the nuScenes and SemanticKITTI datasets demonstrate that our method excels in its simplicity and effectiveness and achieves the best performance in detecting VRU for autonomous driving and road safety.

\bibliographystyle{IEEEtran}
\bibliography{mybibtex}

\end{document}